\documentclass{article}
\usepackage{spconf,amsmath,graphicx}
\usepackage[utf8]{inputenc} 
\usepackage[T1]{fontenc}    
\usepackage{hyperref}       
\usepackage{url}            
\usepackage{booktabs}       
\usepackage{amsfonts}       
\usepackage{nicefrac}       
\usepackage{microtype}      
\usepackage{xcolor}         
\usepackage{graphicx}
\usepackage{subcaption}
\usepackage[numbers]{natbib}
\usepackage[shortlabels]{enumitem}

\title{Neural Video Compression using 2D Gaussian Splatting}
\name{Lakshya Gupta \thanks{*Work done during internship at AMD.} $^{*\ddagger}$  \quad Imran N. Junejo $^\dagger$}
\address{ $^\ddagger$University of Toronto, Canada. \quad $^\dagger$Advanced Micro Devices (AMD), Canada. \\
\texttt{ \{lgupta@cs.toronto.edu, imran.junejo@amd.com\} }}
\begin{document}
%
\maketitle
\begin{abstract}
	The computer vision and image processing research community has been involved in standardizing video data communications for the past many decades, leading to standards such as AVC, HEVC, VVC, AV1, AV2, etc. However, recent groundbreaking works have focused on employing deep learning-based techniques to replace the traditional video codec pipeline to a greater affect. Neural video codecs (NVC) create an end-to-end ML-based solution that does not rely on any handcrafted features (motion or edge-based) and have the ability to learn content-aware compression strategies, offering better adaptability and higher compression efficiency than traditional methods. This holds a great potential not only for hardware design, but also for various video streaming platforms and applications, especially video conferencing applications such as MS-Teams or Zoom that have found extensive usage in classrooms and workplaces. However, their high computational demands currently limit their use in real-time applications like video conferencing. To address this, we propose a region-of-interest (ROI) based neural video compression model that leverages 2D Gaussian Splatting. Unlike traditional codecs, 2D Gaussian Splatting is capable of real-time decoding and can be optimized using fewer data points, requiring only thousands of Gaussians for decent quality outputs as opposed to millions in 3D scenes. In this work, we designed a video pipeline that speeds up the encoding time of the previous Gaussian splatting-based image codec by $88\%$ by using a content-aware initialization strategy paired with a novel Gaussian inter-frame redundancy-reduction mechanism, enabling Gaussian splatting to be used for a video-codec solution, the first of its kind solution in this neural video codec space. 
\end{abstract}
\begin{keywords}
Gaussian Splatting, Neural Video Encoding, Deep Learning
\end{keywords}

\section{Introduction}
The rapid expansion of video-based applications, from teleconferencing to live streaming, has intensified the demand for compression technologies that balance visual quality, bandwidth efficiency, and real-time performance. Traditional video-codecs like H.264 (AVC) and H2.65 (HEVC) rely on handcrafted algorithms for motion compensation and entropy coding, but their rigid architectures struggle to adapt to the dynamic demands of modern high-resolution video \cite{rippel2019learned}. Neural video-codecs (NVCs) have emerged as a promising alternative, leveraging end-to-end machine learning to optimize compression holistically. However, their computational complexity often renders them impractical for latency-sensitive applications, where sub-100ms processing is critical.

A key limitation of existing implicit neural representation (INR) NVC frameworks, such as coordinate-based MLPs or radiance fields, is that they introduce significant encoding/decoding overhead. In contrast, our work introduces 2D Gaussian Splatting as a novel video representation. Unlike implicit methods, Gaussian Splatting provides explicit, pixel-like representations that are inherently compatible with traditional system architectures, enabling efficient GPU/NPU utilization. Each Gaussian can model multiple pixels through its spatial overlap, drastically reducing the number of primitives required to represent a frame. This explicit structure not only improves interpretability but also allows parameters (e.g., position, covariance, opacity) to be dynamically optimized during compression.

To further accelerate encoding, we propose a region-of-interest (ROI)-centric initialization strategy, where Gaussians are prioritized around human subjects in video-conferencing scenarios. By focusing computational resources on semantically critical and meaningful regions, we minimize redundant processing while maintaining perceptual quality. Additionally, our pipeline leverages temporal coherence between frames, reusing Gaussian parameters from previous frames to reduce bitrate demands.

\section{Literature Review}
Recently, researchers have turned their attention to neural video compression. Past studies can broadly be categorized into two main categories: \textit{Autoencoder-style video codecs}  and \textit{MLP-based Implicit Neural Representation} video codecs. 

\textbf{Autoencoder-style video codecs} form a foundational pillar of this evolution, aiming to replicate stages of traditional compression pipelines such as motion estimation, compensation, and residual coding through neural network architectures. Deep Video Compression (DVC) \cite{dvc} marked the inception of this paradigm, employing optical flow networks to predict frame motion and encode residuals. Subsequent works improved this framework by introducing advanced motion prediction techniques, such as scale-space optical flow estimation \cite{scale-space} and recurrent Autoencoders \cite{recurrent-autoencoders}, which enhanced redundancy reduction and coding efficiency. However, these methods primarily operated in the pixel domain, where the predicted frame served as a limited context for compression. The DCVC \cite{dcvc} family addressed this limitation by transitioning from residual coding to conditional coding in the feature domain, enabling richer context learning and improved rate-distortion performance.

Despite their advancements, Autoencoder-based codecs face inherent challenges. Their reliance on extensive parameterization introduces biases tied to the training dataset, complicating generalization. Additionally, the computational complexity of these models, characterized by millions of parameters and high processing demands, limits their adoption in real-time applications. These limitations have driven researchers to explore alternative approaches to neural video compression.

\textbf{Implicit Neural Representations (INRs)} represent a different direction in this space, where videos are encoded as continuous functions using compact neural networks. Pioneering works such as DeepSDF \cite{deepsdf} and NeRF \cite{nerf} showcased the potential of INRs in applications like 3D modeling and image synthesis, leveraging neural networks to parameterize signals in a compact and expressive manner. Whereas INRs offer flexibility and continuous signal representation, they struggle with spectral bias, limiting their ability to capture high-frequency details. Innovations like SIREN’s \cite{siren} periodic activation functions and NeRF’s positional encoding mitigated some of these issues, enhancing representation capacity. For image and video compression, INR approaches treat frames as functions to be learned, storing the network parameters as the compressed representation. NeRV \cite{nerv} introduced the first image-based implicit representations, combining convolutional layers with INRs to accelerate training and inference. Video encoding in NeRV involves fitting a neural network to frames, and the decoding is a straightforward feed-forward process. Initially, the network undergoes training to minimize distortion loss, followed by procedures aimed at diminishing its size, such as converting its weights to an 8-bit floating point format, in addition to quantization and pruning techniques.

Despite these advances, INRs suffer from inefficiencies in decoding and struggle to handle high-resolution video data effectively due to architectural constraints. INRs also face challenges in their ability to discern crucial information necessary for video representation. Although existing approaches strive to improve training and compression methods, there is a continuing need to improve the representation capacity of the network itself. The encoding times for INRs are also very high, usually in the order of 1e-3 FPS, which limits their use in real-time applications. This is because a lot of optimization steps are spent on over-fitting the neural network on a given image or video.

The challenges of both Autoencoder-based and INR-based codecs highlight gaps in neural video compression research. Autoencoders have biases linked to training dataset with generalization challenges, while INRs lack architectural efficiency and struggle to represent high-resolution data with low distortion. To address these limitations, our research explores explicit neural representations using Gaussian Splatting \cite{gsplat}. This method replaces the implicit representation of neural networks with explicit entities, i.e., 2D Gaussians, that directly encode video data. By associating Gaussian parameters such as position, size, and color with specific regions of a video frame, Gaussian Splatting eliminates the reliance on black-box neural models, enabling controlled initialization and scalable representation. This method effectively addresses the expressiveness limitations of INRs while bridging the gap between their computationally intensive encoding processes and the real-time performance of traditional codecs like x.264 \cite{h264}.

\section{Methodology}

Gaussian Splatting (GS) \cite{gsplat} has recently gained tremendous traction as a promising paradigm for 3D view synthesis. With explicit 3D Gaussian representations and differentiable tile-based rasterization, GS not only brings unprecedented control and editability, but also facilitates high-quality and real-time rendering in 3D scene reconstruction. Despite its success in 3D scenarios, the application of GS in video representation remains unexplored. 

In this work, we extend \texttt{GaussianImage} \cite{gimage}, where the authors adapted Gaussian Splatting for 2D image representation,  leveraging the strengths of GS in highly parallelized workflow and real-time rendering to outperform INR-based methods in terms of training efficiency and decoding speed. 

\subsection{2D Gaussian Formation} 
In our framework, the image representation unit is a 2D Gaussian. The basic 2D Gaussian is described by its position $\boldsymbol{\mu} \in \mathbb{R}^2$, 2D covariance matrix $\boldsymbol{\Sigma} \in \mathbb{R}^{2 \times 2}$ and color coefficients $\boldsymbol{c} \in \mathbb{R}^3$. Note that the covariance matrix $\boldsymbol{\Sigma}$ of a Gaussian distribution requires a positive semi-definite. Typically, it is difficult to constrain the learnable parameters using gradient descent to generate such valid matrices. To avoid producing invalid matrices during training, we choose to optimize the factorized form of the covariance matrix. We use Cholesky factorization \cite{cholesky} for decomposition, which breaks down $\boldsymbol{\Sigma}$ into the product of a lower triangular matrix $\boldsymbol{L} \in \mathbb{R}^{2 \times 2}$ and its conjugate transpose $\boldsymbol{L}^T$:

\begin{equation}
	\boldsymbol{\Sigma} = \boldsymbol{L} \boldsymbol{L}^T.
\end{equation}
where we use a Cholesky vector $\boldsymbol{l} = \{l_1, l_2, l_3\}$ to represent the lower triangular elements in matrix $\boldsymbol{L}$. When compared with 3D Gaussians having $59$ learnable parameters, \texttt{GaussianImage}'s (and by extension, our) 2D Gaussian formulation only requires $8$ parameters, making it more lightweight and suitable for image and video representation.

\noindent \textbf{Accumulated Blending-Based Rasterization: }
During the rasterization phase, 3D GS first forms a sorted list of Gaussians $\mathcal{N}$ based on projected depth information. Since the acquisition of depth information involves viewing transformation, it requires us to know the intrinsic and extrinsic parameters of the camera in advance. However, it is difficult for natural individual video to access the detailed camera parameters. In this case, retaining the $\alpha$-blending of the 3D GS without depth cues would result in arbitrary blending sequences, compromising the rendering quality. 

To overcome these limitations, an accumulated summation mechanism is used to fully utilize the potential of the 2D Gaussian representation. Hence, a simplified blending equation is used where we eliminate time-consuming $\alpha$-blending by skipping the tedious sequential calculation of the accumulated transparency $T_n$: 

\begin{equation}
	C_i = \sum_{n \in \mathcal{N}} c'_n \cdot \exp(-\sigma_n),
\end{equation}

\begin{figure*}[h] 
	\centering 
	\includegraphics[width=0.85\textwidth]{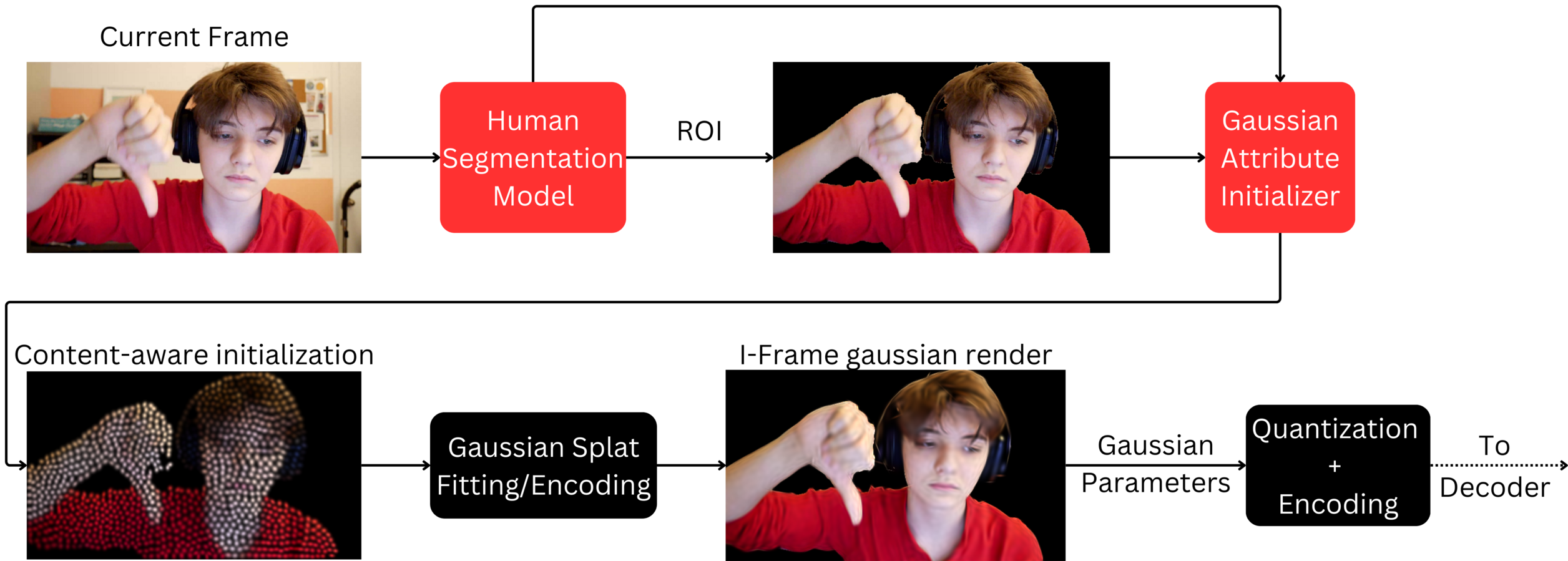} 
	\caption{Pipeline overview of our I-frame representation: region of interest extracted from input frame, passed through our novel Gaussian initializer to get content-aware initialization. This initial Gaussian state goes through optimization steps to get final render, followed by compression. Figure also shows previous work's random initialization}
	\label{fig:1} 
\end{figure*}

\subsection{Frame Representation} 
The proposed codec contains two types of frames: independently coded frames (\textbf{I-frames}) that reduce spatial redundancy and frames that reference a previous frame (\textbf{P-frames}) to reduce temporal redundancy. The codec determines whether to generate an I-frame or a P-frame based on a user-defined Group of Pictures (GoP) sequence length. For example, a GoP length of 4 results in an I-frame every four P-frames. \\

\noindent \textbf{{A. I-Frame Representation Using Content-Aware Initialization Strategy:}} In the proposed codec, this involves optimizing a set of $\mathcal{N}$ 2D Gaussians to fit the frame. A significant latency bottleneck in previous \texttt{GaussianImage} codecs was the random initialization of Gaussian attributes, requiring 13 seconds per frame to reach a target PSNR of 30 on our hardware.

\begin{figure}[h]
	\centering
	\begin{tabular}{ccc}    		
		\hspace{-10pt} \includegraphics[width=0.5\columnwidth]{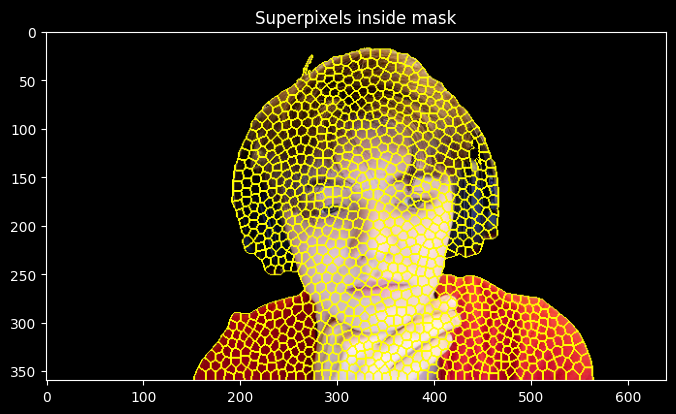} & \hspace{-10pt} \includegraphics[width=0.5\columnwidth]{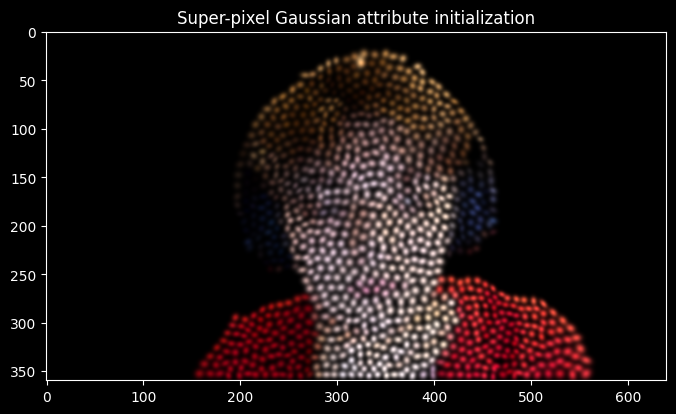} 
		\\
		(a) & (b)  \\    		
	\end{tabular}        
	\caption{Our novel Gaussian attribute initialization strategy: (a) K-Means clustered superpixel segments, (b) Gaussian initialization computed from superpixel segments.}
	\label{fig:2}
\end{figure}

To address this, we introduce a novel initialization strategy by leveraging region-of-interest (ROI) processing. A human matting segmentation model \cite{matting} is used to identify a ROI, defined as the human mask in video conferencing scenarios. This masked region is then segmented into $\mathcal{N}$ superpixel segments using a K-means clustering-based superpixel algorithm \cite{superpixel}. Figure \ref{fig:1} illustrates the pipeline for I-frame representation, while Figure \ref{fig:2} visualizes the superpixel segmentation and Gaussian initialization process.

This segmentation aligns naturally with 2D Gaussian Splatting, as each Gaussian ellipse encompasses multiple pixels. By initializing Gaussian parameters based on superpixel characteristics, the codec achieves content-aware initialization. Specifically, the Gaussian means ($\boldsymbol{\mu}$) are computed as the center of each superpixel, the Cholesky decomposition ($\boldsymbol{L}$) is derived from the covariance matrix of each segment, and the RGB colors ($\boldsymbol{c}$) are calculated as the average pixel color within the segment. This targeted initialization ensures the Gaussians begin close to their optimal state, significantly reducing the number of optimization steps required during Gaussian fitting. Consequently, the proposed approach not only accelerates encoding, but also maintains high visual quality, making it particularly effective for ROI-centric applications. \\

\noindent \textbf{B. Alternative I-Frame Initialization Strategy:}  We explored an alternate approach using a regressor: taking the masked image as input, a regressor is designed to predict parameters for N Gaussians that could effectively represent the input frame. Extensive experimentation was conducted to optimize this neural network-based approach. These experiments included: (1) testing various CNN backbones such as ResNet-50, VGG-19, and ConvNeXt; (2) diversifying and augmenting the dataset through pre-processing techniques; (3) exploring multiple loss functions; (4) employing different activation functions to constrain intermediate representations within bounded spaces; (5) incorporating architectural modifications, such as removing the average pooling layer to preserve spatial information and adding more skip connections to enhance gradient flow and enable information access closer to the prediction head; and (6) dividing the regression task across three specialized neural networks, each responsible for predicting specific Gaussian attributes, to reduce the multitasking burden on a single model.

Among these efforts, the most promising results were achieved using a ResNet-50-based architecture optimized for fast inference. However, the experiments revealed a critical challenge: the unordered Gaussian correspondence problem. Specifically, Gaussian indices lacked consistent associations across frames; for instance, a Gaussian at index $1$ might correspond to a person’s face in one frame but to the chest in another. Simple row-major or column-major sorting only partly solved this issue, as slight positional differences could still lead to vastly different orderings. This inconsistency made point-to-point loss functions like L2 unsuitable, as they failed to establish meaningful associations between specific output neurons and input features.

The Chamfer Distance Loss emerged as an effective solution for addressing this problem by measuring the similarity between two sets of point clouds and establishing one-to-one correspondence. Although this approach performed exceptionally well for predicting Gaussian means and reasonably well for covariance matrices, it struggled to generate accurate RGB values. Consequently, when overlapping ellipses were blended together, they often failed to faithfully reconstruct the input frame. This limitation highlighted the inherent challenges of relying solely on neural network regressors for initializing Gaussian attributes.
\\
\\

\begin{figure*}[t] 
	\centering 
	\includegraphics[width=1.0\textwidth]{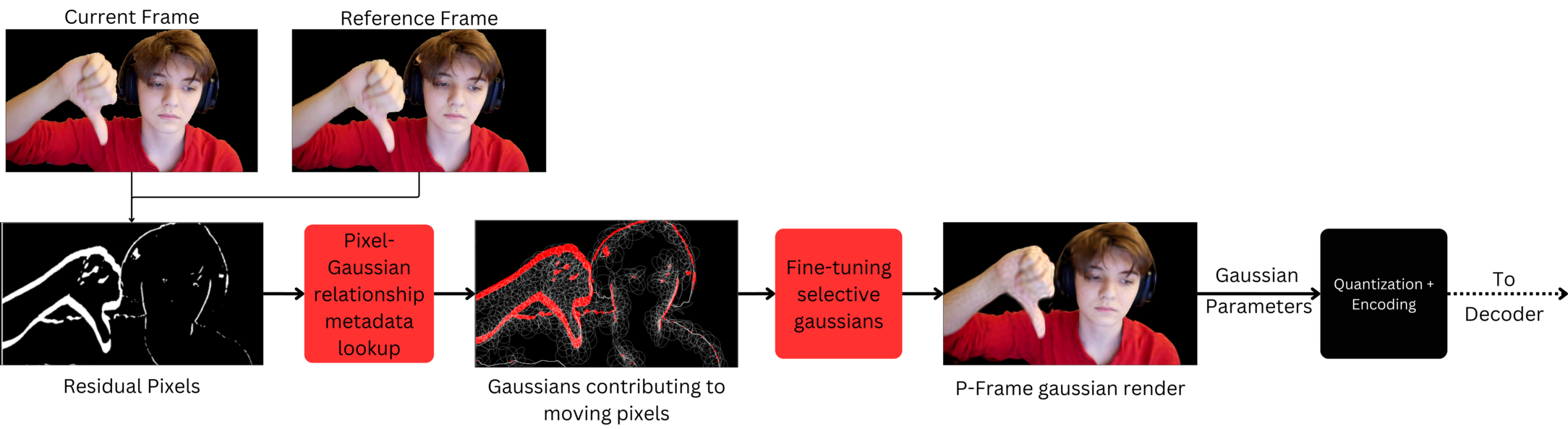} 
	\caption{P-Frame representation workflow: residual pixels between two frames used to compute Gaussians influencing those pixels, which are then selectively optimized followed by compression}
	\label{fig:3} 
\end{figure*}

\noindent \textbf{C. P-Frame Representation Using Selective Gaussian Optimization:} In our proposed codec, we introduce a metadata-based approach to optimize the handling of P-frames. For each frame, metadata is created to capture the mapping between pixels and Gaussians. The pipeline overview of this representation is demonstrated in Figure \ref{fig:3}.

To identify regions of significant change, we compute the residual between the current frame and the reference frame, isolating pixels that exceed a user-defined change threshold. For these identified pixels, the metadata from the reference frame is utilized to locate all Gaussians that influence the changing pixels. Instead of optimizing all Gaussians, only the selected Gaussians associated with the changing pixels are fine-tuned in the current frame.

This selective fine-tuning approach accelerates convergence, as the Gaussians inherited from the reference frame are already closely aligned with the content of the current frame, given the typically minor differences between consecutive frames. Compared to the I-frame approach, this method significantly reduces the number of optimization steps required, thereby decreasing the overall encoding time and transmitted bitstream while maintaining good video quality.

\subsection{Compression Pipeline}

\noindent \textbf{A. Quantization-Aware Training vs. Post-Training Quantization}    
In contrast to previous work in \texttt{GaussianImage} that focused on quantization-aware fine-tuning after overfitting 2D Gaussians, we introduce a more versatile and an efficient approach to quantization. Recognizing the importance of fast encoding, our codec supports both Quantization-Aware Training (QAT) and Post-Training Quantization (PTQ) schemes, enabling flexibility based on the desired trade-off between encoding speed and quality.

For QAT, quantization is integrated directly into the Gaussian optimization process. This involves incorporating a differentiable quantization loss during the Gaussians fitting process. The quantization parameters are iteratively refined during each optimization step to minimize image quality degradation after quantization. Although QAT achieves the highest quality retention post-quantization, it introduces significant computational overhead due to the additional quantization operations performed during each encoding step.

In addition to QAT, our codec also supports PTQ, offering a faster alternative for scenarios requiring low-latency encoding. In PTQ, the Gaussian fitting process is carried out without considering quantization during optimization. Instead, after the Gaussians are fit to the frame, post-training fine-tuning is applied to determine the quantization parameters using a representative dataset, such as a video conferencing dataset. This fine-tuning is only performed during the model's training stage and does not impact the encoding process during inference. As a result, PTQ avoids the time overhead associated with QAT and is well-suited for applications prioritizing rapid encoding, making it the preferred quantization strategy in such cases. 
Our work adopts the quantization strategy from \texttt{GaussianImage} \cite{gimage} and we apply distinct quantization strategies to the three attributes of the 2D Gaussians: mean, cholesky and color. \\

\noindent \textbf{B. Entropy Coding}

In our video codec, we have options for adopting both vanilla entropy coding and bits-back coding \cite{bitsback}. For the later, we take a different approach compared to the prior work in \texttt{GaussianImage}, which relied on partial bits-back coding  for each image. In \texttt{GaussianImage}, the first \(K\) Gaussians in a frame were encoded using vanilla entropy coding, while the remaining \((N-K)\) Gaussians were encoded using bits-back coding. However, as \texttt{GaussianImage} was designed for image codecs, it could not exploit inter-frame dependencies.

To leverage the temporal structure of video data, we encode every I-frame (first frame in a Group of Pictures)  using vanilla entropy coding for all \(N\) Gaussians. For subsequent P-frames, we selectively optimize and entropy code only a subset of Gaussians. On average, \(\frac{N}{20}\) Gaussians are communicated per P-frame. For these frames, we apply bits-back coding, achieving significant bitrate savings.

The total bits required for the entire video sequence are given by: 
\[
\begin{aligned}
	\mathrm{Total\ Bits} &= \mathrm{I\text{-}Frame\ Bits} + \mathrm{P\text{-}Frame\ Bits} \\
	&\quad - \mathrm{Total\ Savings\ for\ P\text{-}Frames}.
\end{aligned}
\]

\noindent \textbf{Savings per P-Frame}:

For P-frames, bits-back coding saves:
\begin{equation}
	S_{\mathrm{P}} = \log\left(\left(\frac{N}{20}\right)!\right) - \log\left(\frac{N}{20}\right).
\end{equation}

The total savings for all P-frames is:
\begin{equation}
	S_{\mathrm{total}} = \left(F - \frac{F}{K}\right) \cdot S_{\mathrm{P}}.
\end{equation}
where \( F \) is the total number of frames,  
\( K \) is the interval for I-frames (e.g., GOP length), and  
\( N \) is the number of Gaussians per frame.

By periodically applying vanilla entropy coding for I-frames and exploiting inter-frame redundancies using bits-back coding for P-frames, our codec achieves a further reduction in bitrate compared to \texttt{GaussianImage} and vanilla entropy coding. Despite its theoretical efficacy, bits-back coding may not align with the objective of developing an ultra-fast codec due to its slow processing latency. Consequently, we leave this part as a preliminary proof of concept on the best rate-distortion performance our codec can achieve. 

\section{Results and Discussions}

\noindent \textbf{Dataset.} Since Our focus is on real-time applications like video conferencing, we test on Microsoft's Video Conferencing Dataset (VCD) \cite{vcd}. VCD consists of $160$ 1080p talking-head video sequences using mutually exclusive subjects and environments. It is organized in four scenarios, each 40 sequences. We downscale the videos to 360p for the current work.

\noindent \textbf{Hardware Configuration.} All the results below have been tested on an Nvidia RTX3090 Ti with AMD Ryzen Threadripper 3960X. 

\subsection{Encoding time improvement}

\begin{figure}[t] 
	\hspace{-10pt}
	\includegraphics[width=.50\textwidth]{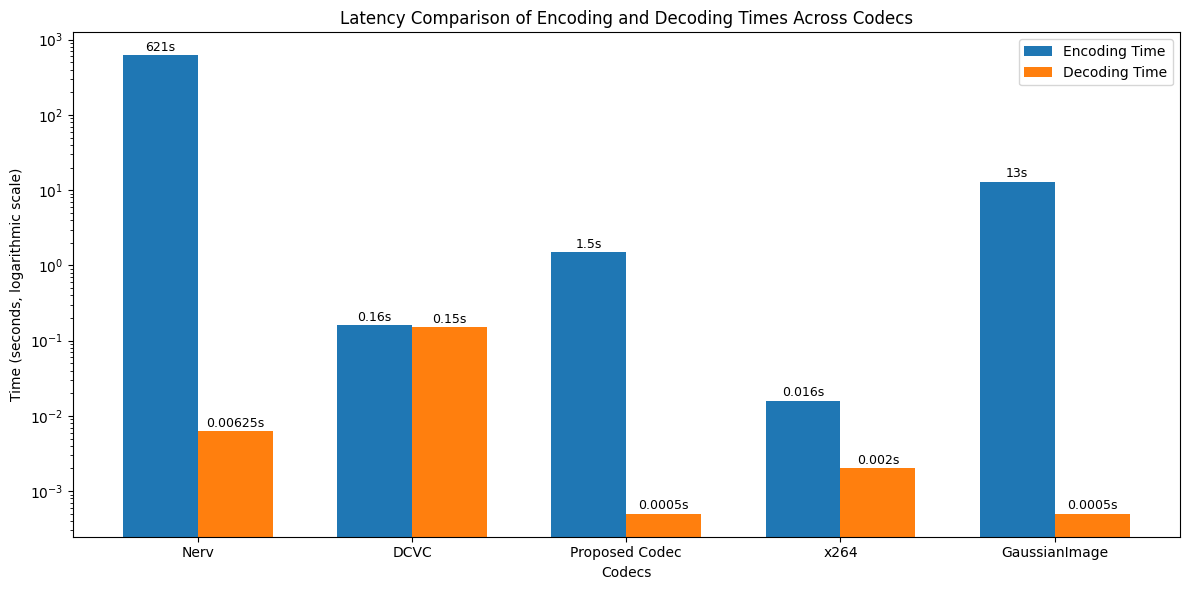} 
	\caption{Encoding and decoding latency comparison of different codecs on VCD dataset}
	\label{fig:4} 
\end{figure}

Figure \ref{fig:4} presents a comparison of encoding and decoding latencies of the proposed codec against several popular codecs, including state-of-the-art autoencoder-based codecs, implicit neural representation (INR)-based codecs, traditional codecs, and prior Gaussian Splatting-based codecs. The results highlight the significant reduction in encoding time achieved by our method. Specifically, the encoding time for I-frames is reduced from $13$ seconds per frame in previous Gaussian Splatting based image codecs (\texttt{GaussianImage}) to just $1.5$ seconds per frame, enabled by our content-aware initialization strategy and dynamic learning rate scheduler. Table \ref{tab:encoding_comparison} shows the impact of our initialization approach compared to \texttt{GaussianImage}. For P-frames, the encoding time is further reduced to $1$ second per frame due to the elimination of initialization overhead, leveraging metadata from the reference frame. These latencies for the proposed codec correspond to $1000$ encoding optimization steps. The proposed codec also delivers the faster decoding performance among the other types of codecs. While INR-based codecs, such as NeRV, achieve real-time decoding, their encoding times are prohibitively high. In contrast, our codec not only bridges the gap in encoding efficiency but also provides ultrafast real-time decoding, making it highly effective for practical video applications. 

\begin{table}[h!]
	\centering
	\small
	\renewcommand{\arraystretch}{0.95} 
	\begin{tabular}{@{}p{3.1cm} p{2.0cm} p{1.8cm}@{}}
		\toprule
		\multicolumn{1}{c}{\textbf{For 1000 Gaussians}} & 
		\multicolumn{1}{c}{\parbox{2.0cm}{\raggedright \textbf{Super-pixel \\ Initializer}}} & 
		\multicolumn{1}{c}{\textbf{GaussianImage}} \\ \midrule
		Iterations for 30 PSNR & \textbf{750} & 8733 \\
		Initialization latency & 100 ms & 15 ms \\
		Total time to 30 PSNR & \textbf{1.5 s} & 13.1 s \\ \bottomrule
	\end{tabular}
	\caption{Comparison of encoding time between super-pixel initialization and \texttt{GaussianImage}.}
	\label{tab:encoding_comparison}
\end{table}

\subsection{Superpixel initialization vs Gaussian Attribute Regressor}
Table \ref{tab:init_comparison} highlights the two initialization strategies explored in this study. Superpixel initialization outperforms random initialization by providing a more informed starting point, resulting in faster convergence to the target video quality. This advantage stems from the neural network regressor’s inability to accurately model the complex blending operations required for Gaussian RGB attributes, despite its effectiveness in capturing Gaussian means and covariances. As the regressor is inherently parallelizable and optimized for GPU acceleration, enhancing its design in future work could significantly accelerate the encoding process. 

\begin{table}[h!]
	\centering
	\small
	\renewcommand{\arraystretch}{0.95} 
	\begin{tabular}{@{}p{3.8cm} p{1.8cm} p{1.8cm}@{}}
		\toprule
		& \multicolumn{1}{l}{\parbox{1.8cm}{\raggedright \textbf{Regressor \\ Initialization}}} 
		& \multicolumn{1}{l}{\parbox{1.8cm}{\raggedright \textbf{Super-pixel \\ Initialization}}} \\ \midrule
		Avg iterations to 30 PSNR & 4000 & \textbf{750} \\
		Avg PSNR at 1000 iterations  & 20.69 & \textbf{30.6} \\
		Initialization latency  & \textbf{6 ms} & 100 ms \\
		Total time to 30 PSNR  & 6 s & \textbf{1.5 s} \\ \bottomrule
	\end{tabular}
	\caption{Comparison of neural network regressor and super-pixel initialization strategies.}
	\label{tab:init_comparison}
\end{table}

\subsection{P-Frame Selective Optimization bitrate savings} 

\begin{figure}[!t] 
	\centering 
	\includegraphics[width=1.0\columnwidth]{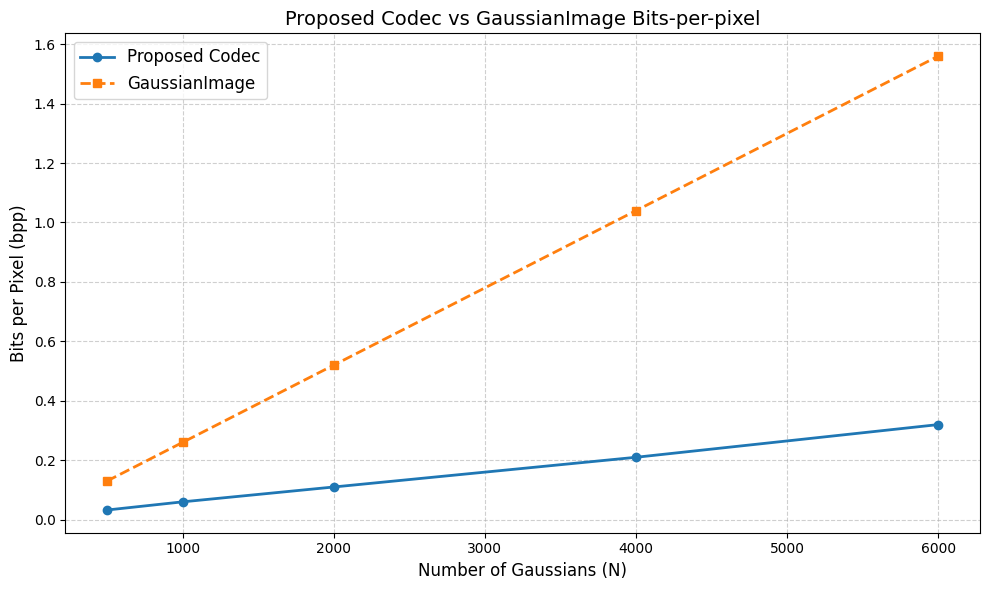} 
	\caption{Bits-per-pixel comparison of our codec with previous GS codec}
	\label{bpp_compare} 
\end{figure}

Selective optimization of Gaussians from the reference frame significantly reduces the bits required for transmission. Figure \ref{bpp_compare} compares the average bits per pixel (bpp) on the VCD dataset for our proposed codec against the previous Gaussian Splatting image codec, \texttt{GaussianImage}. Although \texttt{GaussianImage} is primarily designed for image compression and not for videos, this comparison provides valuable insights into the efficiency of our system in leveraging temporal redundancy for video compression. As the number of Gaussians increases, our proposed codec consistently achieves substantially lower bpp compared to \texttt{GaussianImage}, demonstrating the effectiveness of our approach in reducing the bitstream size. Notably, the proposed codec reduces the average bitstream size by 78\%, highlighting its capability for efficient representation and compression in video applications. This significant reduction underscores the impact of temporal optimization on reducing transmission costs while maintaining competitive quality. 

\subsection{Rate-distortion comparison to other neural codecs} 

\begin{figure*}[!t]
	\centering        
	\begin{tabular}{ccc}    		
		\includegraphics[width=0.3\textwidth]{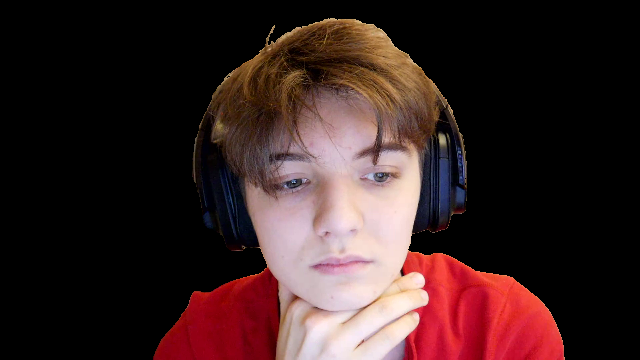}  & \includegraphics[width=0.3\textwidth]{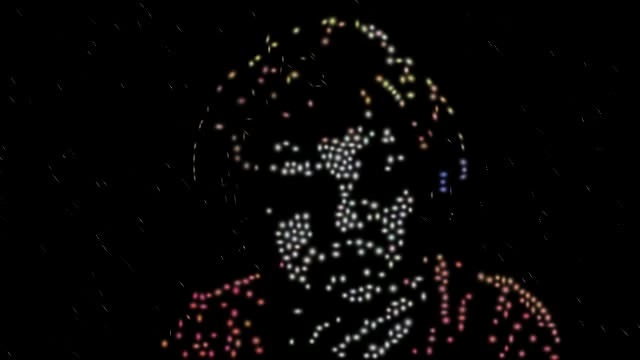} & \includegraphics[width=0.3\textwidth]{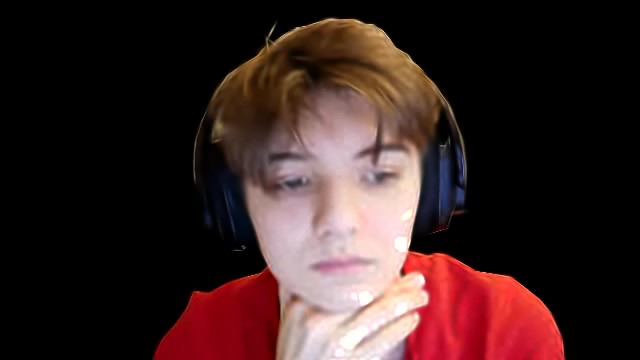}
		\\
		(a) & (b) & (c)  \\    		
	\end{tabular}        
	\caption{Comparison of codec's visual output with previous GS codec on fast setting (1000 iterations). (a) Original frame. (b) Previous GS codec. (c) Our GS codec.}
	\label{fig:5}
\end{figure*}

\begin{figure}[t] 
	\hspace{-10pt}
	\includegraphics[width=1\columnwidth]{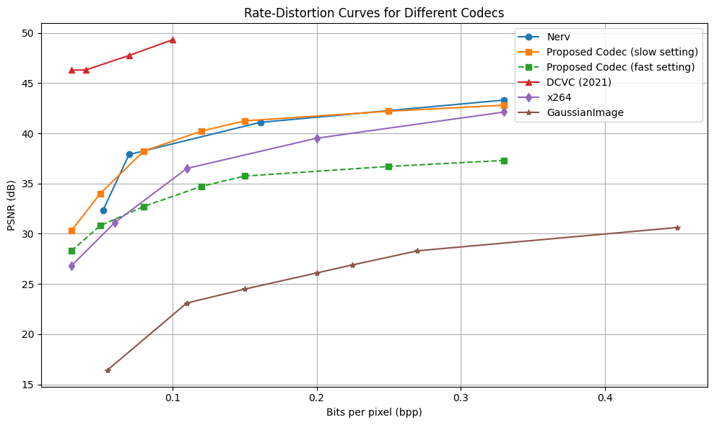} 
	\caption{Rate-distortion curves of different codecs; low bpp and high PSNR is ideal.}
	\label{fig:6} 
\end{figure}

Figure \ref{fig:6} depicts the rate-distortion (R-D) performance of the proposed codec compared to other codecs, where the quality metric, Peak Signal-to-Noise Ratio (PSNR), is plotted against the compression factor (bits per pixel, bpp). A lower bpp corresponds to a smaller bitstream size, while a higher PSNR indicates better reconstruction quality. The proposed codec's R-D performance is presented for two configurations: a ``fast" setting with $1,000$ encoding iterations and a ``slow" setting with $10,000$ encoding iterations. These settings provide a trade-off between encoding speed and quality, offering user-configurable flexibility based on application requirements. The ``fast" setting emphasizes low latency, whereas the ``slow" setting demonstrates the upper bounds of achievable quality for the given codec architecture.

The results indicate that under the ``slow" setting, the proposed codec achieves performance comparable to NeRV, an implicit neural representation (INR)-based codec, while outperforming traditional codecs such as x264 in terms of quality. Furthermore, in the ``fast" setting, the codec achieves higher PSNR at low bpp compared to x264, demonstrating its efficiency in resource-constrained scenarios. In both configurations, the proposed codec consistently outperforms the previous Gaussian Splatting-based image codec, highlighting its advancements in both quality and compression efficiency. This dual-configurability not only reinforces the codec's versatility across diverse applications but also showcases its capability to bridge the gap between real-time encoding and high-quality compression. Figure \ref{fig:5} compares our codec output to the previous GS codec under fast encoding. Our initializer achieves significantly better image quality within $1000$ iterations, while \texttt{GaussianImage} remains in early convergence with lower fidelity.

\subsection{Post-Training-Quantization vs Quantization-Aware-Training}

\begin{figure*}[h]
	\centering
	\begin{subfigure}[b]{1.0\columnwidth} 
		\centering
		\includegraphics[width=0.9\columnwidth]{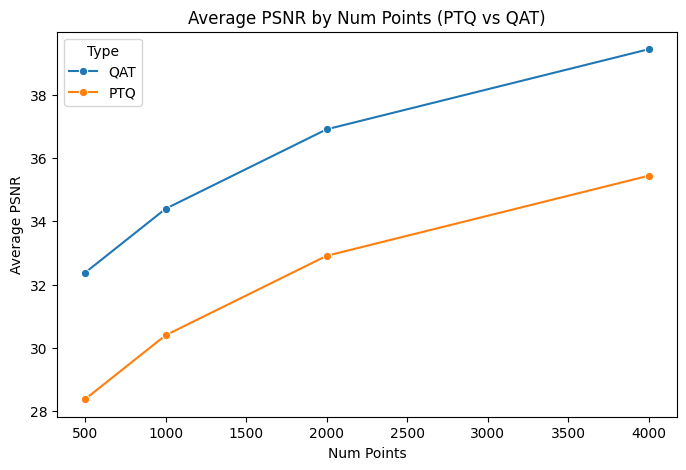} 
		\caption{\small Impact with varying number of Gaussians} 
		\label{fig:ptq_qat_a}
	\end{subfigure}
	\hfill
	\begin{subfigure}[b]{1.0\columnwidth} 
		\centering
		\includegraphics[width=0.9\columnwidth]{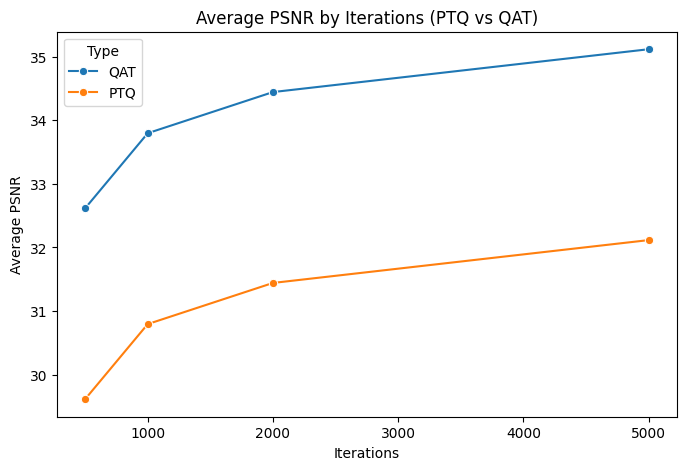} 
		\caption{\small Impact with varying number of iterations} 
		\label{fig:ptq_qat_b}
	\end{subfigure}
	\vspace{-10pt} 
	\caption{PTQ vs QAT impact on video quality (PSNR)}
	\label{fig:ptq_qat}
\end{figure*}

Fig. \ref{fig:ptq_qat} compares PTQ and QAT quantization schemes in terms of PSNR and latency. While PTQ offers lower quality than QAT, it significantly boosts encoding speed (Table \ref{tab:ptq_qat}). We recommend PTQ for fast encoding scenarios like video conferencing and QAT for offline streaming. Additionally, Fig. \ref{fig:ptq_qat} highlights that the number of Gaussians have a greater impact on video quality than the number of iterations. 

\begin{table}[t]
	\centering
	\small
	\renewcommand{\arraystretch}{1.2} 
	\setlength{\tabcolsep}{4pt} 
	\begin{tabular}{|p{3.2cm}|p{1.4cm}|p{1.4cm}|}
		\hline
		\multicolumn{1}{|c|}{\textbf{For 1000 GS Fitting Iterations}} & \textbf{PTQ} & \textbf{QAT} \\
		\hline
		Quantization latency & 3.5 ms & 3.5 s \\
		\hline
	\end{tabular}
	\caption{Comparison of quantization latency between PTQ and QAT for 1000 Gaussian Splatting (GS) fitting iterations.}
	\label{tab:ptq_qat}
\end{table}

\subsection{Online video streaming usability comparison}

For real-time video streaming applications, a fundamental requirement of the underlying codec is the ability to transmit bit information on a frame-by-frame basis. While this requirement is inherently supported by traditional codecs and Autoencoder-based codecs, it is not feasible for INR-based codecs like NeRV. This limitation arises because INR-based approaches first overfit a neural network to the entire video and subsequently transmit the network's weights as a whole, making them unsuitable for live video transmission. Consequently, INR-based codecs are more aligned with offline streaming scenarios rather than real-time applications.

In contrast, our proposed explicit neural representation-based video codec supports frame-by-frame bit transmission, meeting the foundational criteria for real-time streaming. By leveraging Gaussian splatting, our codec achieves significant improvements over previous Gaussian Splatting-based methods (i.e. \texttt{GaussianImage}), increasing encoding speeds from $0.07$ FPS to $0.67$ FPS while maintaining real-time decoding speeds. Although Autoencoder-based codecs like DCVC-DC currently offer much higher encoding speeds (15 FPS), our approach demonstrates the feasibility of explicit neural representation codecs for real-time applications. While not yet optimized for video conferencing use cases, this work showcases the potential of Gaussian Splatting-based codecs to bridge the gap toward real-time video streaming.

\section{Challenges and Future work}

Looking at key challenges and opportunities for improvement, the primary issue lies in the visual artifacts observed when reconstructed frames are stitched into a video, particularly in ``fast" encoding settings. These artifacts, referred to as Gaussian motion artifacts, arise because Gaussians representing the same region in consecutive frames undergo independent optimization, resulting in slight parameter differences. When stitched into a video, these differences manifest as a subtle noise-like motion over moving regions, such as a person’s body. This is further exacerbated by incomplete Gaussian convergence in fast encoding, leading to blurry low-frequency details.

Another challenge is the less-than-expected reduction in P-frame encoding time. Although P-frames encode faster ($1$ second per frame) than I-frames ($1.5$ seconds), their starting point, optimized Gaussians from the reference frame, suggests they should converge in significantly fewer iterations. Projections estimated P-frame encoding times at 100-200 milliseconds ($60$-$120$ iterations), yet the current implementation requires considerably more, limiting its efficiency gains.

Additionally, while the proposed codec reduces encoding time by more than 8× compared to prior Gaussian Splatting-based codecs, achieving $0.66$ FPS, it is still not real-time. Furthermore, the current focus on encoding only the region of interest (ROI) limits the codec’s applicability. Future work should explore hybrid encoding strategies, where the background is encoded using traditional codecs at high compression rates, enabling Gaussian Splatting-based encoding for the ROI to maintain high quality while achieving real-time performance. Improvements to P-frame optimization, such as leveraging optical flow directly on 2D Gaussians, could also address encoding time bottlenecks, though this remains a challenging problem due to the non-one-to-one Gaussian-pixel mapping.

By addressing these challenges, the proposed codec could become a practical, real-time solution for video compression with superior quality in ROI-focused applications.

\section{Conclusions} 

In this work, we proposed a fast neural video compression model built upon the real-time rendering framework of 2D Gaussian Splatting (i.e. \texttt{GaussianImage}). By utilizing explicit representations through Gaussian Splatting, our approach addresses key limitations of Autoencoder-based methods, such as generalization challenges, and the inefficiencies of implicit neural representations (INRs), which struggle with varying resolution data and slow encoding speeds. Our model achieved a significant reduction in encoding latency—over 8x faster compared to previous Gaussian Splatting (GS)-based state-of-the-art and INR approaches—through a novel initialization strategy and selective Gaussian optimization. Our codec is also among the very few that can do frame-by-frame bistream transmission and provide realtime decoding, making it eligible for realtime video conferencing applications. Despite these advancements, the encoding latency remains below real-time performance. These findings highlight the potential of Gaussian Splatting for video representation and point to promising directions for future research focused on achieving real-time encoding speeds.


\bibliographystyle{IEEEbib}
\bibliography{main_bib}

\end{document}